\documentclass[12pt]{article}
\usepackage{amsmath}
\usepackage{graphicx,psfrag,epsf}
\usepackage{enumerate}
\usepackage{natbib}
\usepackage{url} %
\usepackage{xcolor}

\newcommand{\blind}{0}

\usepackage{algorithm}
\usepackage{algpseudocode}
\usepackage{amsfonts}
\usepackage{subcaption}
\usepackage{graphicx}

\usepackage{hyperref}

\begin{document}
\def\spacingset#1{\renewcommand{\baselinestretch}%
{#1}\small\normalsize} \spacingset{1}

\newcommand{\xm}[1]{\textcolor{red}{XM: #1}}
\newcommand{\at}[1]{\textcolor{red}{#1}}
\newcommand{\sm}[1]{\textcolor{blue}{SM: #1}}

\if0\blind
{
  \title{\bf Cox-Hawkes: doubly stochastic spatiotemporal Poisson processes}
  \author{Xenia Miscouridou$^1$\footnote{Correspondence to x.miscouridou@imperial.ac.uk},  Samir Bhatt$^{1,2}$, George Mohler$^3$,\\ Seth Flaxman$^4$$^\dagger$, Swapnil Mishra$^2$\footnote{Equal contribution}
  }
  \date{
    $^1$ Imperial College London
    $^2$ University of Copenhagen
    $^3$ Boston College\\
    $^4$ University of Oxford 
    }
  
  \maketitle

} \fi

\if1\blind
{
  \bigskip
  \bigskip
  \bigskip
  \begin{center}
    {\LARGE\bf Cox-Hawkes: doubly stochastic spatiotemporal Poisson processes}
\end{center}
  \medskip
} \fi

\bigskip
\begin{abstract}%
Hawkes processes are point process models that have been used to capture self-excitatory behaviour in social interactions, neural activity, earthquakes and viral epidemics. They can model the occurrence of the times and locations of events. Here we develop a new class of spatiotemporal Hawkes processes that can capture both triggering and clustering behaviour and we provide an efficient method for performing inference.
We use a log-Gaussian Cox process (LGCP) as prior for the background rate of the Hawkes process which gives arbitrary flexibility to capture a wide range of underlying background effects (for infectious diseases these are called endemic effects). The Hawkes process and LGCP are computationally expensive due to the former having a likelihood with quadratic complexity in the number of observations and the latter involving inversion of the precision matrix which is cubic in observations. Here we propose a novel approach to perform MCMC sampling for our Hawkes process with LGCP background, using pre-trained Gaussian Process generators which provide direct and cheap access to samples during inference. We show the efficacy and flexibility of our approach in experiments on simulated data and use our methods to uncover the trends in a dataset of reported crimes in the US.%
\end{abstract}
\noindent%
\vfill
\newpage
\spacingset{1.45} %
\section{Introduction}
\label{sec:intro}

Hawkes processes are a class of point processes that can model self or mutual excitation between events, in which the occurrence of one event triggers additional events, for example: a violent event in one geographical area on a given day encourages another violent event in an area nearby the next day. A unique feature of Hawkes processes is their ability to model exogenous and endogenous "causes" of events. An exogenous cause happens by the external addition of a event, while endogenous events are self-excited from previous events by a triggering kernel. An example of the difference between these two mechanisms is in disease transmission - an exogenous event could be a zoonosis event such as the transmission of Influenza from birds, while endogenous events are subsequent human to human transmission. Due to their flexibility and mathematical tractability, Hawkes processes have been extensively used in the literature in a series of applications. They have modelled among others, neural activity~\citep{Linderman2014b}, earthquakes~\citep{Ogata1988}, violence~\citep{Flaxman2015,Holbrook2021} and social interactions~\citep{Miscouridou2018}. 

The majority of applied research on Hawkes processes focuses on the purely temporal settings where events occur and are subsequently triggered only in time. However, many practical problems require the inclusion of a spatial dimension. This inclusion is motivated by several factors, first, natural phenomena that self-excite tend to do so both spatial and temporally e.g. infectious diseases, crime or diffusion over a network. Second, natural processes tend to cluster closely in space and time~\citep{Tobler1970}. Third, in parametric formulations residual variation persists and this is often structured in both space and time~\citep{Diggle1998}. A wide body of research exists in modelling spatial phenomena ranging from Kriging~\citep{Matheron1962} to model based estimates~\citep{Diggle1998} using Gaussian processes. In the more general Gaussian process, which provides a prior function class, spatial phenomena are modelled through a mean function and a covariance function that allows control over the degree of clustering as well as the smoothness of the underlying functions. Specifically for applications for spatial point patterns, an elegant formulation using log-Gaussian Cox processes (LGCP),~\cite{Moller1998} is commonly used~\citep{Diggle2013}. LGCPs can capture complex spatial structure but at a fundamental level are unequipped with a mechanism to model self-excitement. When examining the processes' endogenous and exogenous drivers, the lack of a self-exciting mechanism can potentially lead to spurious scientific conclusions even if prediction accuracy is high. For example, appealing again to the Influenza example, only modelling the distribution of cases using an LGCP will ignore the complex interplay of zoonosis events and secondary transmission events, both of which require different policy actions.

The inclusion of space has a long history via the Hawkes process triggering mechanism - fistly modelled using the Epidemic Type Aftershock Sequence (ETAS) kernel~\cite{Ogata1988} but many subsequent approaches now exist. However, to our knowledge, very few approaches consider spatial and temporal events in \emph{both} the exogenous and endogenous Hawkes process mechanisms - that is where events can occur in space and time, and then these events trigger new events also in space and time. Many mechanisms have been proposed for space-time triggering kernels~\cite{Reinhart2018}, but it is not clear nor straightforward how to also allow for exogenous space-time events simultaneously. In the vast majority of previous applications, exogenous events occur at a constant rate in both space and time or with highly specific forms that depend on the setting e.g. periodic functions for seasonal malaria data~\citep{Unwin2021}. Some studies do provide nonparametric approaches for the background rate:~\cite{Lewis2011} provide an estimation procedure for the background and kernel of the Hawkes process when no parametric form is assumed for either of the two.~\cite{Miscouridou2018} use a nonparametric prior based on completely random measures to construct the discrete background rate for the Hawkes processes that build directed networks. Other recent approaches use neural networks to estimate the rate~\citep{Omi2019}. However, these nonparametric approaches do not provide a compelling stochastic mechanism, such as LGCPs, that yield a generative process. 

Here we propose a novel time space approach that combines Hawkes processes~\citep{Hawkes1971} with log-Gaussian Cox processes~\citep{Moller1998,Diggle2013}. This synthesis allows us, for the first time, to have a exogenous background intensity process with self-excitation that is stochastic and able to vary in both space and time. We provide a suite of new methods for simulation and computationally tractable inference. Our methods leverage modern computational techniques that are scalable and can efficiently learn complex spatiotemporal data. We apply our approach on both simulated and real data. Our novel addition of an LGCP prior in both space and time is accompanied with new computational challenges: a Hawkes process is quadratic in complexity due to a double summation in the likelihood, and LGCPs incur cubic complexity from matrix inversions. To ensure our approach is scalable and still competitive with standard Hawkes processes we utilize a recently developed Gaussian process approximation~\citep{Mishra2020,Semenova2022} that obliviates the need for repeated matrix inversions. Our work represents a step towards more general, scalable, point process framework that encodes more flexible and plausible mechanisms to represent natural and physical phenomena.

\subsubsection*{Our contributions}
A summary of the contributions of our work is: (i) We provide a novel model formulation for a highly flexible self-exciting process that can capture endogenous and exogenous events in both space and time. Our utilization of LCGPs for the exogenous background rate is extremely flexible and follows from the current state-of-the-art in spatial statistics~\citep{Diggle2013}, (ii) in contrast to previous work (e.g.~\cite{Flaxman2015}) our framework admits a generative model that can produce stochastic realizations at an arbitrary set of locations. We provide a novel algorithm to sample from this generative process, (iii) we offer an efficient Bayesian inference approach that ensures our more flexible model is still as scalable as standard Hawkes processes and straightforward to implement computationally, (iv) our framework is directly applicable to numerous spatiotemporal problems where there are both endogenous and exogenous causes e.g. for natural or social phenomena such as crime, diseases, environment, or human behaviour.

\section{Related methods}	

As mentioned before, modelling space through Hawkes processes was fist used with the Epidemic Type Aftershock Sequence (ETAS) kernel~\citep{Ogata1988} and other approaches followed some of which exist in~\cite{Reinhart2018}. For modelling spatial point patterns without self-excitation, log-Gaussian Cox processes (LGCP)~\cite{Moller1998} provide an elegant approach as explained in~\cite{Diggle2013}.

\cite{Reinhart2018} provide an overview on spatiotemporal Hawkes processes explaining various options for the form of the intensity, the kernels and the corresponding simulating algorithm. However, the case of an LGCP background is not discussed in the review or elsewhere.
 
Our approach is the first to use an LGCP to capture the background underlying effects (these are called endemic effects in infectious disease modelling but here we will use this term broadly for other applications too) and can model the exact spatial and time locations.
 
\cite{Flaxman2015} aim to understand whether gun violence in Chicago is contagious or merely clusters in space and time. To this end, they use a spatiotemporal Hawkes model and a space-time test to distinguish between the two. The model uses a kernel density estimator for the background (endemic) effects and a kernel for the epidemic events that is separable in space and time. Their model has a different construction as it does not admit a generative procedure since the background rate is estimated using kernel density estimators.

Similarly to~\cite{Flaxman2015},~\cite{Holbrook2021} build a scalable inference algorithm for parametric spatiotemporal self-exciting  processes. The proposed model is the one of~\cite{Flaxman2015} which is based on a Gaussian kernel smoother for the background. The main contribution is to overcome the bottleneck of the quadratic computational complexity of such a point process. The authors develop a high-performance computing statistical framework to do Bayesian analysis with Metropolis-Hastings using contemporary hardware. They apply it on a gunfire dataset which covers a larger dataset and more fine-grained than the one in~\cite{Flaxman2015}. 
 
 The combination of a Hawkes process with an LGCP is found in~\cite{Linderman2015} where the authors propose a purely temporal multivariate Hawkes process with LGCP in the background with the goal to infer a latent network structure given observed sequences of events. This approach is based on~\cite{Linderman2014} but in discrete time and with an improved inference scheme based in mini batches. However both of these two have different scope to our work and work only with univariate time data.
 
 Finally,~\cite{Mohler2013} develops a purely temporal Hawkes process model with LGCP background for count (aggregated) events.~\cite{Mohler2013} builds a Metropolis adjusted Langevin algorithm for estimation and uses the algorithms to disentangle the sources of clustering in crime and security data. We are instead interested in modelling and predicting exact event times and locations. %

\section{Model}
\label{sec:model}
\subsection{Hawkes process definition}
A Hawkes process is an inhomogenerous Poisson point process defined in terms of a counting measure and an intensity function or rate. Hawkes processes were originally proposed by~\cite{Hawkes1971} as temporal point processes. The intensity is conditional on the history of the process such that the current rate of events depends on previous events. We focus on self-exciting Hawkes processes, in which historic events encourage the appearance of future events.

Here we develop spatiotemporal self-exciting processes which can predict the rate of events happening at specific locations and times. For a spatiotemporal Hawkes process on the domain $\mathcal{X}\times [0,T)$, for $\mathcal{X}\subset \mathbb{R}^d$ we denote the counting measure of the process by $N$ %
and the conditional intensity by $\lambda$. The definition of the inhomogeneous point process intensity is as below.
For $\mathbf{s} \in \mathcal{X}\subset \mathbb{R}^d$ (generally $d$ here represents Euclidean or Cartesian coordinates) and $t\in [0,T)$ 
\begin{align}
\lambda(t,\mathbf{s})&=\lim_{\Delta t,\Delta s\rightarrow 0} \frac{\mathbb{E}[N[\left(t,t+\Delta t) \times B(\mathbf{s},\Delta \mathbf{s})\right]|\mathcal{H}_t]}{\Delta t\times |{B(\mathbf{s},\Delta \mathbf{s})}| } 
\label{eq:def_intensity}
\end{align}
where $\mathcal{H}_t$ denotes the history of all events of the process up to time t, $N(A)$ is the counting measure of events over the set $A \subset \mathcal{X} \times [0,T)$ and $|B(\mathbf{s},\mathbf{s})|$ is the Lebesgue measure of the ball $B(\mathbf{s},\mathbf{s})$ with radius $\mathbf{s}>0$.
Note that the spatial locations can be univariate, referring for example to regions or countries, or bivariate such as geographical coordinates of longitude and latitude or even multivariate depending on the context.

\subsection{Hawkes process intensity}
\label{sec:intensity}
The conditional intensity defined as in Eq~\eqref{eq:def_intensity} admits the form
\begin{align}
    \lambda(t,\mathbf{s}|\mathcal{H}_t)&=\mu(t,\mathbf{s}) + \sum_{i:t_i<t} g\left(t-t_i,\mathbf{s}-\mathbf{s}_i\right),
    \label{eq:hawkes}
\end{align}
where $(t_1,t_2,\dots,t_n )$ denotes the ordered sequence of the times of the observed events and $({s}_1,{s}_2,\dots,{s}_n )$ their corresponding spatial locations. 
Events arise either from the background rate $\mu(t,\mathbf{s})$
(exogenous or non excitation effects) or from the triggering function/kernel $g$ (endogenous or self-excitation effects). Both $g$ and $\mu$ must be nonnegative.

$g$ can take a parametric form or be estimated using full nonparametric assumptions, as done for example in~\cite{Rousseau2020}. $g$ can also  be separable (additive or multiplicative) or non-separable, in space and time. Here we take a separable form of a product of an exponential kernel in time and a Gaussian kernel in space but more complex forms are possible. For any $t>0$ and $\mathbf{s}\in \mathcal{X}\subset \mathbb{R}^d$ the self-exciting part of the rate is given by 
\begin{align}
\label{eq:triggering_kernel}
g(t,s)&=\alpha \beta \exp\left(-\beta t\right) \frac{1}{\sqrt{2\pi|\Sigma|}}\exp\left(-{\mathbf{s}^T\Sigma^{-1}\mathbf{\mathbf{s}}}\right),
\end{align}
where $\alpha>0, \beta >0$ and $\Sigma$ a semi positive definite matrix. The condition for stationarity implies $\alpha < 1$, with an expected total cluster size of
$\frac{1}{1 - \alpha}$.

The triggering function $g$, centered at the triggering
event, is the intensity function for the offspring process. Properly normalized, it induces a probability distribution for the location and times of the offspring events. The cluster process representation of the Hawkes process will prove crucial to the efficient simulation of
self-exciting processes which we give in section~\ref{subsec:simulation}.
For the temporal part we use the widely used exponential kernel, originally proposed by~\cite{Hawkes1971}, giving exponential decay which is suitable for the applications we are interested in. For the spatial part, we use a Gaussian kernel which is suitable for modelling spatial locations and is again a popular choice for triggering the spatial locations used by other authors in literature such as~\citep{Flaxman2015}.

An important property of the process is the branching ratio $b=\int _{\mathcal{X}}\int_0^\infty g(dt,d\mathbf{s})$ which represents the ratio of the number of total offspring to the size of the entire family. We require that $b \in (0, 1)$ which ensures that the process is stationary which is one of the basic assumptions for the properties of the Hawkes process to be well defined. For $b=0$ we have a Cox process where as for $b\geq 1$ the process explodes (the standard definition of explosion is that $N(t)-N(\mathbf{s})=\infty$ for $t-\mathbf{s}<\infty$). To see how explosion emerges, we refer the reader to section $5.4$, Lemma $2$ of~\cite{Grimmett2001} which give the calculations on the expected number of descendants of one atom. More on the implications of the values of $b$ can be found in~\cite{Asmussen2003}.

In this context, the Hawkes process is stationary when the jump process
$\left\{dN(t): t > 0\right\}$, which takes values in $\{0, 1\}$ is weakly stationary. Therefore the kernel needs to satisfy the condition $\int_{\mathcal{X}}\int_0^\infty g(t,\mathbf{s})dt d\mathbf{s}<1$. Stationarity ensures that cluster sizes are almost surely finite, and that since generation of offspring follows a geometric progression, the expected total cluster size is
$\frac{1}{1 - \int_{\mathcal{X}} \int_0^\infty g(t,\mathbf{s})dt d\mathbf{\mathbf{s}}}$ including the initial background event. %

Given the form of the intensity, past events influence future ones and depending on the form of the triggering function $g$, the process may have short or long term effects. In either case, the kernel is responsible for the self-exciting behaviour of the process.

$\mu(t,\mathbf{s})$ is the background rate of the process. It is a nonnegative function with initial nonzero value that captures the underlying patterns in space and time that encourage the clustering of events in those time and space locations. It often takes the form of a constant for simplicity, or a parametric form such as periodic as assumed in~\cite{Unwin2021} or can even have a nonparametric prior constructed on random measures as in~\citep{Miscouridou2018}. As further explained in more detail below, we assume a log-Gaussian process prior on $\mu(t,\mathbf{s})$. %

\subsection{Latent log Gaussian process for background rate}

We use a latent Gaussian process ($\mathcal{GP}$) to determine the background rate of events in time $t\in R$ and space $\mathbf{s}\in R^d$. This means that the background rate takes the form 
\begin{align}
\mu(t,\mathbf{s})=\exp\left(f\left(t,\mathbf{s}\right)\right)
\end{align}
where $f(t,\mathbf{s})$ is a function realization from a Gaussian process prior in space and time. Formally, a Gaussian process is a collection of random variables, such that any finite collection of them is Gaussian distributed. $\mathcal{GP}$s are a class of Bayesian nonparametric models that define a prior over functions which in our case are functions over time and space. 
Similarly to a probability distribution that describes
random variables which are scalars or vectors (for multivariate distributions), a Gaussian process is distribution over functions and belongs in the family of stochastic processes.

$\mathcal{GP}$s are a powerful tool in machine learning, for learning complex functions with applications in regression and classification problems. We refer the reader to~\cite{Rasmussen2005} for details on Gaussian processes and their properties. 

A Gaussian process on $\mathbb{R}^D$, for any $D>0$ is completely specified by its mean function $\alpha(\cdot)$ and covariance function $k(\cdot,\cdot)$. For $\mathbf{u} \in \mathbb{R}^D$ we will denote a draw from a Gaussian process as $$f(\mathbf{u})\sim \mathcal{GP}\left(\alpha_u, k\left(\mathbf{u},\mathbf{u}'\right)\right).$$

The Gaussian process is centered around its mean function, with the correlation structure (how similar two points are) of the residuals specified via the covariance kernel. Properties of the underlying function space such as smoothness, differentiability and periodicity can be controlled by the choice of kernel. One of the most popular choices of covariance kernel, and the one we choose to introduce the model with, is the Gaussian kernel (also commonly called the squared exponential kernel), defined for $u,u' \in \mathbb{R}^D$ by the covariance function
\begin{align}
\label{eq:covariance_kernel}
   Cov\left(f\left(\mathbf{u}\right),f\left(\mathbf{u}'\right)\right)= k\left(\mathbf{u},\mathbf{u}'\right)=\omega^2\exp\left(-\frac{1}{2l^2}|\mathbf{u}-\mathbf{u}'|^2\right)%
\end{align}
 where $|\mathbf{u}|$ denotes the Euclidean norm, i.e.  it is equal to $|\mathbf{u}|=\sqrt{\sum_i \mathbf{u}_i^2}$ if $u$ is a vector ($D>1$ e.g. the spatial locations) and to the absolute value of $\mathbf{u}$ if $\mathbf{u}$ is a scalar ($D=1$ e.g. timestamps). $\omega^2>0$ defines the kernel?s variance scale and $l>0$
is a length scale parameter that specifies how nearsighted the
correlation between pairs of events is.

The hyperparameters can be varied, thus also known as free parameters. The kernel and mean of the $\mathcal{GP}$ together fully specify the prior distribution over functions.

We will consider an additive separable kernel with a bivariate spatial dimension $\mathbf{s}=(x,y)$ and univariate temporal dimension $t$.  In order to have a nonnegative background we exponentiate the additive kernel. From this kernel specification the background intensity $\mu(t,s)$ follows a log-Gaussian Cox process~\citep{Moller1998, Diggle2013} over space and time
\begin{align}
\label{eq:background}
\mu(t,s)&=\exp\left(f_{s}\left(\mathbf{s}\right)+f_t\left(t\right)\right) \\
f_t &\sim \mathcal{GP}\left(m_t,k_t\right) \nonumber\\
f_{s} &\sim \mathcal{GP}\left(m_{s},k_s\right) ,\nonumber
\end{align}
where $m_t$ and $m_{s}$ are the $\mathcal{GP}$ mean functions and $k_t$, $k_s$ are the kernels defined by %
the hyperparameters $\omega^2_t, \omega^2_s, l_t,l_s$.%

\subsection{Full Model Likelihood}

To model the spatial coordinates $\mathbf{s}=(x,y)$ and time stamps $t$, we use a Hawkes kernel $g_{ts}(t,s)=g_t(t)g_s(s)$ and a log-Gaussian Cox process $\mu(t,s) = \exp\left(f_s(s)\right)\exp\left(f_t(t)\right)$. Without loss of generality we will assume here that the Gaussian processes have zero mean. The joint model we consider is a Hawkes process with composite rate $\lambda(t,x,y)$ which is the sum of the intensities of an LGCP process and a Hawkes process
 \begin{align}
 \label{eq:rate}
 \lambda(t, x,y)&= \exp \left(f_s\left(x,y\right)+ f_t\left(t\right) \right) \nonumber \\&+ \sum_{i:t_i<t} g_t(t-t_i)g_s(x-x_i,y-y_i) \nonumber\\
 &=\exp \left(f_s\left(x,y\right)+ f_t(t) \right)\nonumber \\&+ \sum_{i:t_i<t}\alpha \beta \exp\left(-\beta (t-t_i)\right) \frac{1}{2\pi \sigma_x \sigma_y}\exp\left(-\frac{(x-x_i)^2}{2\sigma^2_x}-\frac{(y-y_i)^2}{2\sigma^2_y}\right).
\end{align}
One could see this as an intercept coming from constant contributions by both the temporal and spatial background processes as these are only identifiable through their sum and not separately. Given a set of observed ordered times $\left(t_1,t_2,\dots,t_n \right) \in [0,T)$ and the corresponding locations $\left(s_1,s_2,\dots,s_n \right) \in \mathcal{X}$, let $D$ denote the full data $D=\{t_i, s_i\}_{i=1}^n$ and $L(D)$ the likelihood. Following~\cite{DaleyVereJones2008} the likelihood is given by
\begin{align}
\label{eq:likelihood}
    L(D)&=\left[\prod_{i=1}^{n} \lambda(t_i,s_i)\right]\nonumber \exp\left(-\int_\mathcal{X} \int_0^T \lambda(t,\mathbf{s})dtd\mathbf{s}\right)\\
    &=\left[\prod_{i=1}^{n} \lambda(t_i,x_i,y_i)\right]  \exp\left(-\int_\mathcal{X} \int_0^T \lambda(t,x,y)dtdxdy\right).
\end{align}
We give below details on how to simulate from the process with the rate defined in Eq~\eqref{eq:rate} and perform inference using the likelihood in Eq~\eqref{eq:likelihood}.

\section{Methods}
\label{sec:methods}

\subsection{Simulation}
\label{subsec:simulation}
By construction our model admits a generative process facilitating simulation. This is an important and nuanced advantage over previous spatiotemporal models~\citep{Flaxman2015,Holbrook2021} which were not fully generative due to a deterministic paramaterization of the exogenous component. Note that the model of~\cite{Mohler2013} does admit a generative model but only for a purely temporal model for aggregated (count) data. In general Hawkes processes can be simulated in two ways: through an intensity based approach or a cluster based approach. We give below Algorithm~\ref{alg:simulation} to simulate from our model via the latter approach, i.e. through simulating the background first and then the generations of subsequent offsprings. %
Note that for the hyperparameters $l_t,l_s,\omega^2_t,\omega^2_s$ one can either fix them to a known value or (hyper)priors on them.
\begin{algorithm}[h!]
\caption{Cluster based generative algorithm for Hawkes process simulation} \label{alg:simulation}
\begin{algorithmic}
\Require Fix $T>0, \mathcal{X}$
\State \textbf{Draw}  $l_t,l_s \sim p^+(\cdot)$
\State \textbf{Draw}  $\omega^2_t,\omega^2_s \sim p^+(\cdot)$
\State \textbf{Draw} $f_t \sim \mathcal{GP}(0,k_t)$, $f_s \sim \mathcal{GP}(0,k_s)$
\State \textbf{Set} $\mu(t,\mathbf{s})=\exp(f_t(t)+f_s(\mathbf{s}))$
\State \textbf{Draw} $N_0\sim Pois\left(\int_{0}^{T}\int_{\mathcal{X}} \mu\left(t,\mathbf{s}\right)\, dt d\mathbf{s}\right)$
\State \textbf{Draw} $\{t,\mathbf{s}\}_{i=1}^{N_0} \sim \mu(t,\mathbf{s})$
\State \textbf{Set} $G_0=\{t,\mathbf{s}\}_{i=1}^{N_0},\, \ell=0$
\While{$G_\ell \neq 0$}
\For{$i=1$ to $N_\ell$}
    \State Simulate $C_i$ the number of offspring of event $i$ events
    \State Simulate $O_i,\dots,O_{C_i}$ the pairs for the inter arrival times and distances of offsprings
    \EndFor
    \State $\ell+=1$
    \State $G_\ell=\{G_{\ell-1}+\bigcup_{i=1}^{N_\ell} O_1, \dots, O_{C_i} \}_{<T}$
\EndWhile
\end{algorithmic}
\Return $\bigcup_{\ell}G_\ell$
\end{algorithm}

We use a clustering approach for simulation which makes use of the branching ratio $b$ (see section~\ref{sec:intensity}) and relies on the following idea: for each immigrant $i$, the times and locations of the first-generation offspring arrivals given the knowledge of the total number of them are each i.i.d. distributed.
We provide the simulation in Algorithm~\ref{alg:simulation}.
As a test check for making sure that our Hawkes process simulations are correct we employ an approximate Kolmogorov-Smirnov type of test adapting Algorithm $7.4.V$ from~\cite{DaleyVereJones2008}.
In Algorithm~\ref{alg:simulation} $p^+$ refer to a probability distribution on the real line.

To simulate from our model proposed above, i.e. a Cox-Hawkes process we need to draw from a $\mathcal{GP}$. Since $\mathcal{GP}$s are infinitely dimensional objects, in order to simulate them we have to resort to finite approximations. The most common approach is to implement them through finitely-dimensional multivariate Gaussian distributions. This is the approach we take as well for simulating our $\mathcal{GP}$s. In order to sample points from the LGCP background  of the process, we draw an (approximate) realization from the Gaussian process prior and then use rejection sampling to sample the exact temporal and spatial locations. %

\subsection{Inference}
Given a set of $n$ observed data $\mathcal{D}=\{t_i,\mathbf{s}_i\}_{i=1}^n$ over a period of $[0,T]$ and a spatial area denoted by $\mathcal{X}$, we are interested in a Bayesian approach to infer the parameters and hyperparameters of the model. Denote by $\theta$ and $\phi$ set of the parameters of the background rate $\mu(t,\mathbf{s})$ and the triggering rate $g(t,\mathbf{s})$ respectively. This gives $\theta=\left(f_t(t),f_s(\mathbf{s})\right)$ %
$\phi=\left(\alpha,\beta,\sigma^2_x,\sigma^2_y\right)$. 

The posterior is then given by
\begin{align}
\label{eq:posterior}
\pi(\phi,\theta|D) &\propto \pi(\theta)\, \times  \pi(\phi) \times L(\mathcal{D})\\ \nonumber
&= \pi(f_t(t))\pi(f_s(\mathbf{s}))\times  \pi(\alpha)\pi(\beta) \pi(\sigma^2_x) \pi(\sigma^2_y) \times  L(\mathcal{D}). \nonumber
\end{align}
where $ L(\mathcal{D})$ is the likelihood defined in Eq~\eqref{eq:likelihood} and with some abuse of notation, we use $\pi$ to denote both prior and posterior for all the parameters. %
The prior for $\pi(\alpha)$, $\pi(\beta)$, $\pi(\sigma^2_x)$ and $\pi(\sigma^2_y)$ is Truncated Normal (restricted on the positive real line) to ensure the positivity of these parameters.
Note that the prior on the functions $f_t$ and $ f_s$ can be further defined by the priors on the hyperparameters %
$l_t\sim \text{InverseGamma}$, $\omega^2_t \sim \text{LogNormal}$ for the temporal process and $ l_s \sim \text{InverseGamma}$, $  \omega^2_s \sim \text{LogNormal}$ for the spatial. The objective is to approximate the total posterior distribution $\pi(\phi, \theta|D)$ using sampling methods. %

A classical Hawkes process has quadratic complexity for computating the likelihood. Only in special cases such as that of a purely temporal exponential kernel the complexity is reduced from quadratic to linearas it admits a recursive construction. See \cite{Dassios2013} for an explanation. Note however we cannot apply this in our case as it does not hold when we add (on top of the temporal exponential kernel) the Gaussian spatial kernel. Inference is in general cumbersome and people tend to either resort to approximations or high performance computing techniques such as \cite{Holbrook2021}. %

A naive formulation of combining log-Gaussian cox processes as the background intensity function in the spatiotemporal Hawkes process will increase the computational complexity for the inference. This happens because in addition to the quadratic complexity arising from the triggering kernel the exogeneous formulation naively introduces a cubic complexity for a lgcp~\citep{Diggle2013}.

We propose to circumvent the computational issues through a reduced rank approximation of a Gaussian process~\citep{Semenova2022} through variational autoencoders (VAE). This approach relies on pre-training a VAE on samples from a Gaussian process to create a reduced rank generative model. Once this VAE is trained, the decoder can be used to generate new samples for Bayesian inference. More specifically, in this framework one should first train a VAE to approximate a class of $\mathcal{GP}$ priors (the class of GP priors learned varies from context to context depending on our prior belief about the problem space) and then utilizes the trained decoder to produce approximate samples from the $\mathcal{GP}$. This step reduces the inference time and complexity as drawing from a standard normal distribution $ z \sim \mathcal{N}(0,\mathbb{I})$ with uncorrelated $z_i$ is much more efficient than drawing from a highly correlated multivariate normal $\mathcal{N} \sim (0,\Sigma)$ with dense $\Sigma$. For more details see section $2.5$ in~\cite{Semenova2022}. Here we will denote this approximation to the Gaussian Process prior by $\tilde \pi $. %
Hence, we obtain %
overall the Bayesian hierarchical model
\begin{align}
\label{eq:post}
\pi(\phi,\theta|D)&\propto \pi(\theta)\, \pi(\phi) L(D) \nonumber \\
&=\pi(f_t(t))\pi(f_s(s))\pi(\alpha)\pi(\beta) \pi(\sigma^2_x) \pi(\sigma^2_y) \nonumber \\
&\approx \tilde \pi\left(f_t(t)\right)\tilde \pi(f_s(s)) \pi(\alpha)\pi(\beta) \pi(\sigma^2_x) \pi(\sigma^2_y).
\end{align}

The code for simulation and inference for this class of models of Cox-Hawkes processes implemented in python and numpyro~\citep{phan2019composable} can be found at
\url{https://anonymous.4open.science/r/Spatiotemporal_Cox-Hawkes}.
\section{Experiments}
\label{sec:experiments}
We demonstrate the applicability of our methods on both simulated and real data. For simulations our goal is twofold: (i) to show that we can accurately estimate the parameters of both the background and self-exciting components thereby showing that we recover the true underlying mechanisms and (ii) to show that our method performs well under model misspecification, thereby showing our model is sufficiently general to be used in real data situations where the true underlying data generating mechanism is unknown. On real settings we apply our methods to gunfire data used in~\cite{Flaxman2015} detected by an acoustic gunshot locator system to uncover the underlying patterns of crime contagion in space and time. We show how our model can be used as an actionable tool by practitioners to understand and measure contagion effects in important settings. 
Note that throughout the following we refer to the model used for simulating data as the true model.
\subsection{Experiment 1: Simulated Data}
\begin{figure}[!tbp]
        \centering
        \includegraphics[width=.7\linewidth, height=.3\textheight]{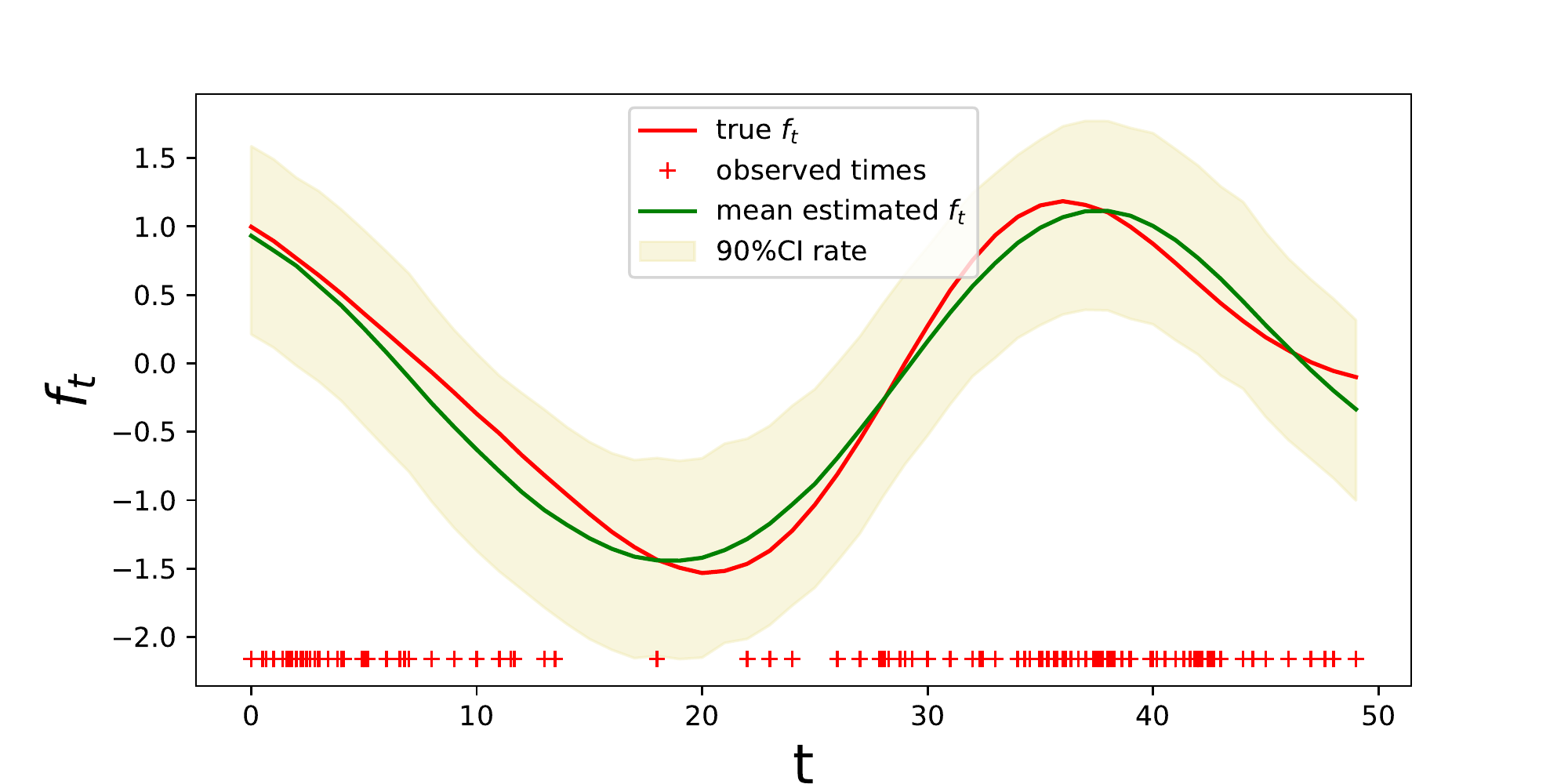}
        \caption{Plot for the temporal Gaussian process $f_t(t)$ on simulated data. The red line is the simulated draw of the Gaussian process, the green line is the mean posterior and the yellow shaded area is the $90\%$  credible interval. The red marks on the x-axis are the exact simulated times from the background process.}
        \label{fig:post_pred_t_simul_A}
\end{figure}
\begin{figure}[!tbp]%
        \centering
        \includegraphics[width=.8\linewidth, height=.35\textheight]{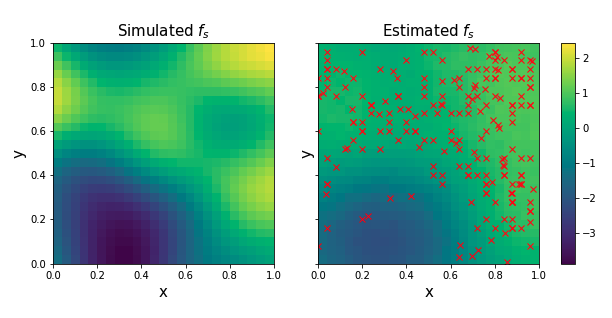}
        \caption{Simulated draw and the posterior predictive distribution for the 2-dimensional spatial Gaussian process. The simulated $f_s(x,y)$ is shown on the left on a regular grid and the mean predictive distribution is shown on the right with the simulated locations in red.}
        \label{fig:post_pred_s_simul_A}
\end{figure}

We simulate data from a Hawkes process with rate as given in Eq~\eqref{eq:rate} on the domain $[0,T]=[0,50], \mathcal{X}=[0,1]\times[0,1]$. 
For the background rate which governs the exogenous events we simulate a realization of the latent (separable) spatiotemporal Gaussian process with covariance kernels defined as in Eq~\eqref{eq:covariance_kernel} using 
$l_t=10,\omega^2_t=1, l_s=0.25, \omega^2_s=1 $. 
The simulated $f_t(t)$ from the temporal Gaussian process can be seen in Figure~\ref{fig:post_pred_t_simul_A} in red and the temporal events drawn from this background are also shown in red on the x-axis. 
The simulated two-dimensional spatial Gaussian process can be seen at the left plot of Figure~\ref{fig:post_pred_s_simul_A}.
Note that we also use an explicit constant intercept of $a_{0}=0.8$ giving an overall background rate of $\exp(a_0+f_t +f_s)$ and in inference we use a Normal prior on it.

For the diffusion effect we use a triggering spatiotemporal kernel of the form in Eq~\eqref{eq:triggering_kernel} with values $\alpha=0.5, \beta=0.7$ for the exponential kernel. For the Gaussian spatial kernel we will assume a common parameter $\sigma^2$ for both $\sigma^2_x$ and $\sigma^2_y$ which we will assume to be $0.5$. This gives a set of around $n=210$ spatiotemporal points $\{ t_i, x_i, y_i\}_{i=1}^n$ of which the ratio of background to offspring events is roughly $1:1$.

\begin{figure}[!tbp]
        \centering
        \includegraphics[width=\linewidth, height=.35\textheight]{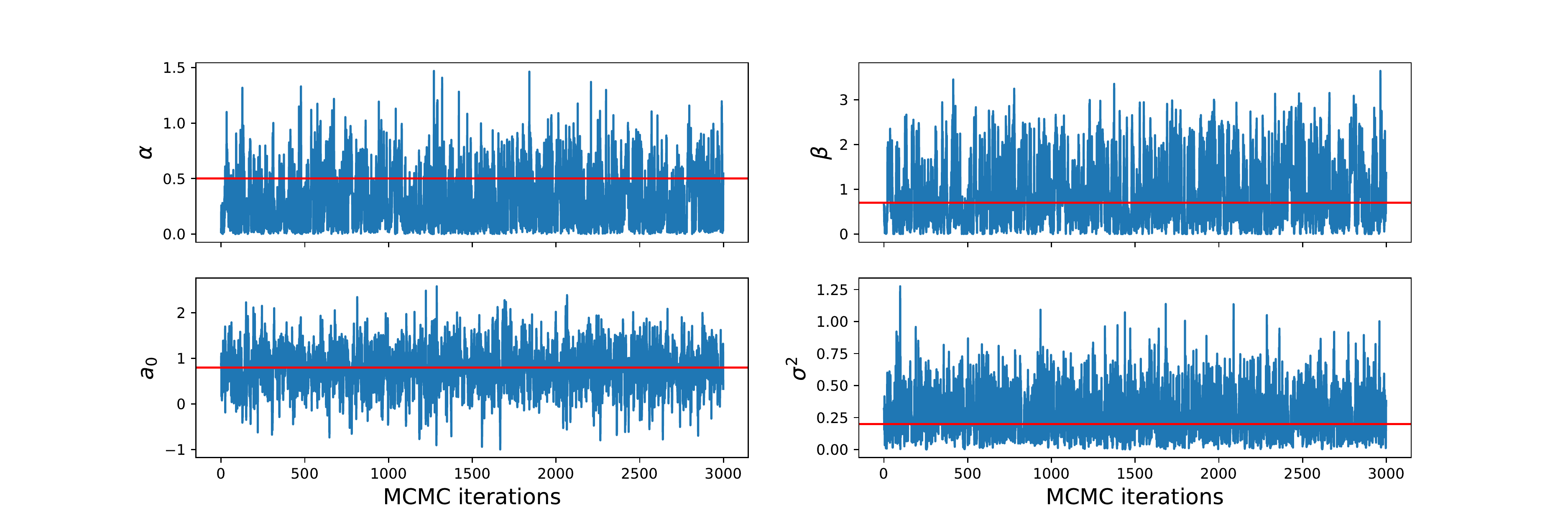}
        \caption{MCMC trace for $a_0,\alpha,\beta, \sigma$ on simulated data where the red line shows the simulated value for the experiments. The samples shown are collected from $3$ chains.}
        \label{fig:trace_simul_A}
\end{figure}

For inference we run {$3$ chains with $1,500$ samples each of which $500$ were discarded as burn in, using a thinning size of $1$.} %
In Figure~\ref{fig:trace_simul_A} we report the trace plots for the parameters $\alpha, \beta, \sigma$ which define the triggering kernel that governs excitation. We also report $a_0$ which we used as the total mean of the latent Gaussian process $\mu(t,x,y)=\exp\left(f_t(t)+a_0+f_s(x,y)\right)$. We use a Normal prior on $a_0$. %
In all cases the simulated values shown in red are within the trace coverage. In our experiments we combine the samples (after removing the warmup iterations) from all the chains. The plots overall show good convergence with good mixing between the chains and no multimodal behaviour. 
Regarding the Gaussian process fitting, we show the posterior predictive plots in Figures~\ref{fig:post_pred_t_simul_A} and ~\ref{fig:post_pred_s_simul_A}.
For the one-dimensional temporal Gaussian process we plot the simulated draw $f_t(t)$ in Figure~\ref{fig:post_pred_t_simul_A} in red. The blue line is the mean posterior of $f_t(t)$ and the yellow shaded area is the $90\%$ credible interval obtained from the posterior predictive distribution. The red dots on the x-axis are the exact simulated time events drawn from the process. The $90\%$ credible interval covers well the simulated function, and the mean posterior predictive is very close to the simulated one, showing good model fit.

For the two-dimensional spatial Gaussian process we plot the simulated draw function $f_s(x,y)$ at the left plot of Figure~\ref{fig:post_pred_s_simul_A}. The the mean posterior predictive distribution is shown in the centre and the mean predictive distribution with the true simulated locations embedded on it is shown on the right. The color scale on the right shows the relative values ranging from dark blue (smallest) to yellow (highest). The simulated and the mean posterior predictive are relatively similarly when compared visually which shows a good fit for the model. 

To quantify convergence we require good $\hat R$ diagnostics. For all our results the $\hat R$ diagnostics returned by the sampler were in $[1,1.002]$ for all the estimated parameters. This, combined with visual inspection of the MCMC trace show good evidence of convergence and good mixing behaviour.%

\subsection{Experiment 2: Model misspecification}
Our second experiment on simulated data compares and contrasts our method (LGCP-Hawkes) to a Hawkes process with constant background and a pure log Gaussian Cox Process. %
The intensity for our LGCP-Hawkes model is Eq~\eqref{eq:rate}, for Hawkes it is Eq~\eqref{eq:hawkes} with constant $\mu(t,\mathbf{s})=\mu$ and for LGCP it is Eq~\eqref{eq:background}.%

We simulate data from these three inhomogeneous point process models and then fit each model on every dataset on a train set and perform prediction on a test set. Note that we also fit under a homogeneous Poisson model as it's the baseline giving the simplest spatiotemporal model that exists.
We show that our model Hawkes-LGCP is a reasonable approach even when there is model mismatch (i.e. when the data are drawn from a pure Hawkes or pure LGCP). It is therefore a good approach to use in real data scenarios when the underling data generating mechanism is unknown.

It is challenging to evaluate the quality of model fit from different point process models, and especially their generalization ability. Reporting a quantity such as the likelihood for test data is not possible as the evaluation depends on the actual events. We therefore adopt a procedure to predict the temporal and spatial locations of future events under the inference model and then compute the combined error between those events and events generated under the true generating model that was used for simulation. This procedure mirrors the properties that practitioners would desire from their model in real world settings.
The distance metric we use is the root mean square error (RMSE) difference between the exact simulated and predicted events. We show below that in a model misspecification scenario, using our proposed model is a good approach as an inference model to be used for prediction purposes. %
The experimental setup is as follow. We simulate $100$ datasets (each of which give on average $300$ events) over a fixed time window and a fixed spatial domain and then do a train-test split. We repeat this using as generating model each of the models.%
  \begin{figure}[!tbp]
        \centering
        \includegraphics[width=1\linewidth, height=.3\textheight]{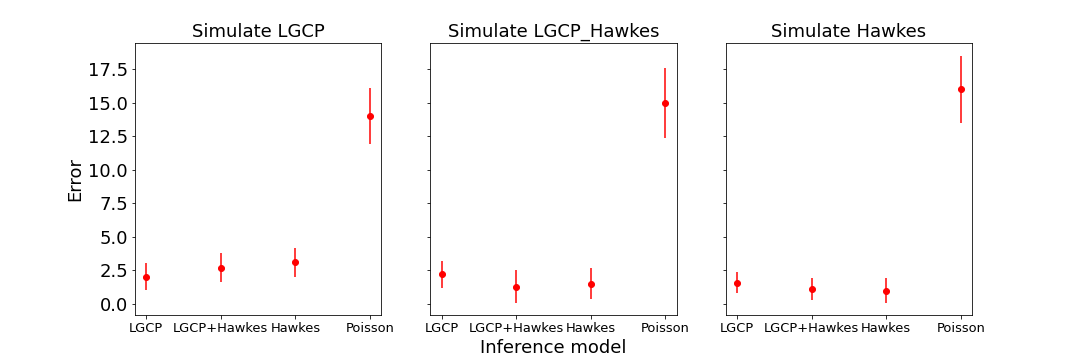}
        \caption{The average RMSE and its standard error reported for the model misspecification experiment. The left plot corresponds to a simulated dataset from an LGCP model, the middle to an LGCP$+$Hawkes and the right from a Hawkes model. In all three cases we perform inference under all LGCP, LGCP$+$Hawkes, Hawkes as well as Poisson (baseline). %
        }
        \label{fig:error}
\end{figure}
For every dataset, we then perform inference under our MCMC scheme under every model. Given the estimated parameters, we predict $200$ times the next $10$ future events which we compare to those of the test set. We compute the error between the true and estimated events across the $200$ predictions and the $100$ simulations. We report the mean and standard error of the RMSE graphically in Figure~\ref{fig:error}. This shows how good each model is in predicting the near future.

As shown in Figure~\ref{fig:error}, and as expected, the error is always lowest when the true model is used for inference, however in all cases the next best model is LGCP-Hawkes although the differences are not always statistically significant. This provides evidence for our model's ability to flexibly capture a wide range of underlying patterns. In all cases the worst model is the Poisson baseline, as its constant intensity in space and time cannot capture the inhomogeneities in the data. These results highlight that when the true data generating process is unknown, which is the default scenario in real world settings, our model is likely to be a robust choice.
\subsection{Experiment 3: Gunshot Dataset}

\begin{figure*}[!tbp]
\centering
\begin{subfigure} {.48\textwidth}
\includegraphics[width=\textwidth,height=0.3\textheight]{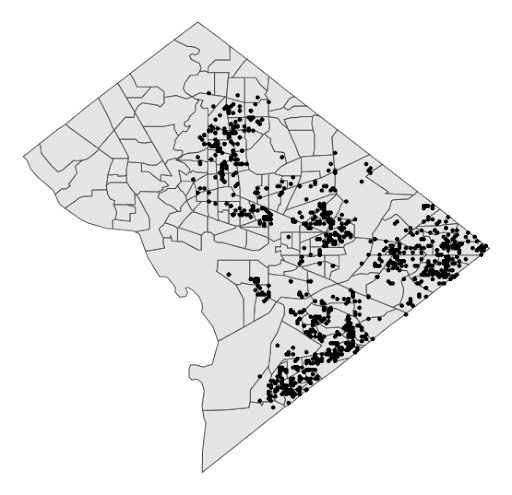}
\caption{}
\end{subfigure}
\begin{subfigure} {.48\textwidth}
\includegraphics[width=\textwidth,height=0.3\textheight]{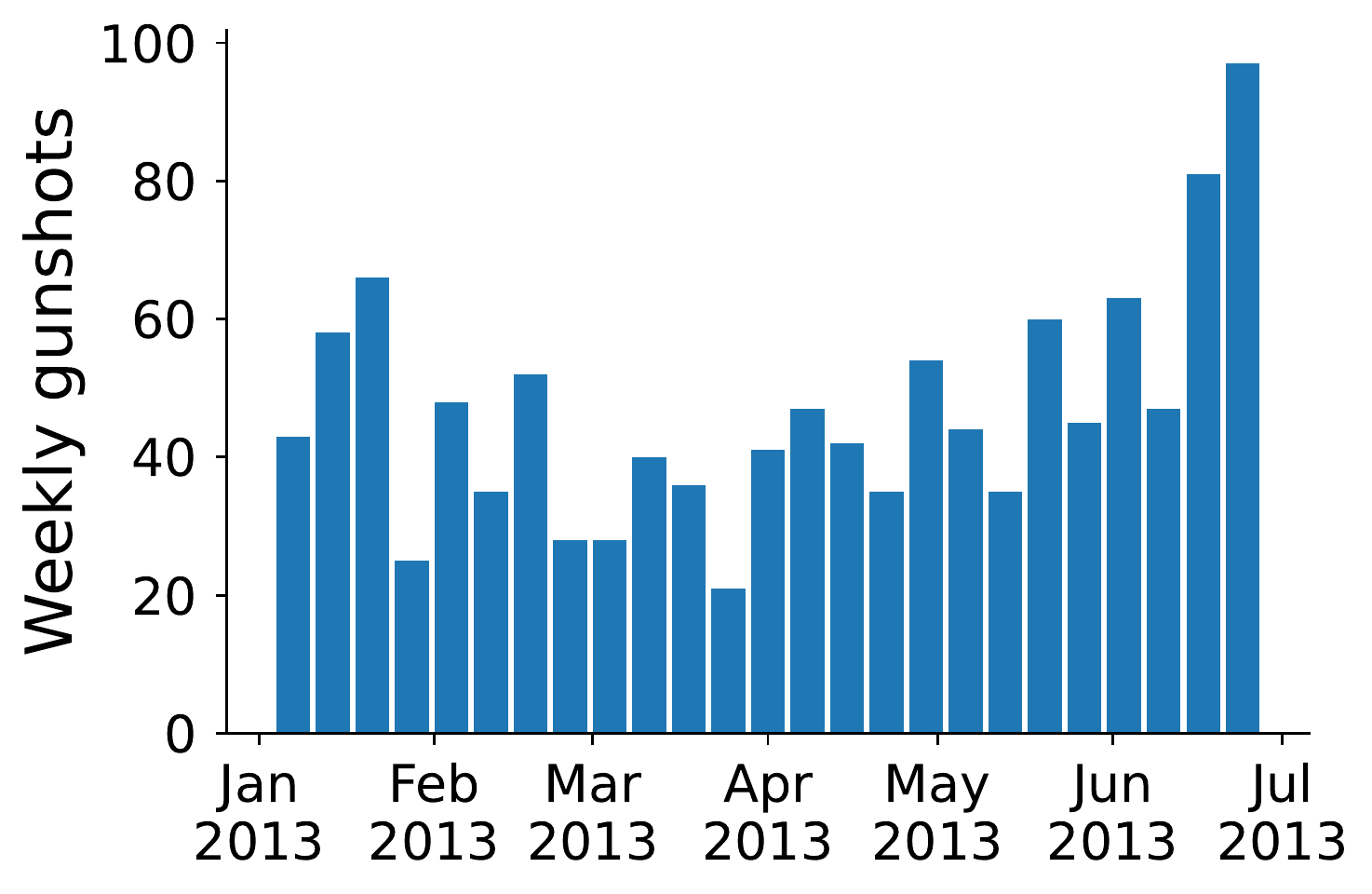}
\caption{}
\end{subfigure}
\caption{
Spatial (a) and temporal (b) distribution of the gunfire data in Washington DC over the year 2013. The spatial locations are the exact geographical coordinates and the temporal locations are shown weekly.}%
\label{fig:empirical_gunshots}
\end{figure*}
We use gunshot data in $2013$ recorded by an acoustic gunshot locator system (AGLS) in Washington DC and follow~\cite{Flaxman2015} for data preprocessing. %
There were 1,171 gunshots recorded in total. Spatial locations were rounded to produce approximately $100$m spatial resolution and $1$ sec temporal resolution. Visualizations of the temporal and spatial distributions of the data are shown in Figure~\ref{fig:empirical_gunshots}(a) and (b).
\begin{figure*}[!tbp]
\centering
\begin{subfigure} {.48\textwidth}
\includegraphics[width=\textwidth,height=0.3\textheight]{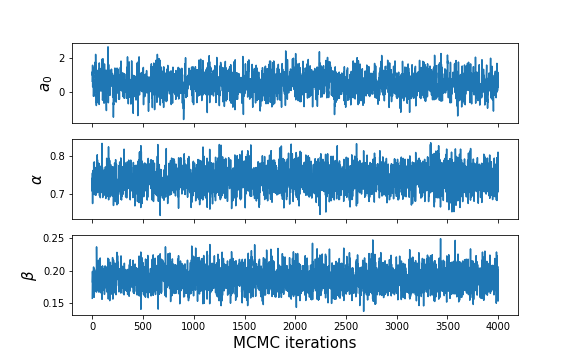}
\caption{}
\end{subfigure}
\begin{subfigure} {.48\textwidth}
\includegraphics[width=\textwidth,height=0.3\textheight]{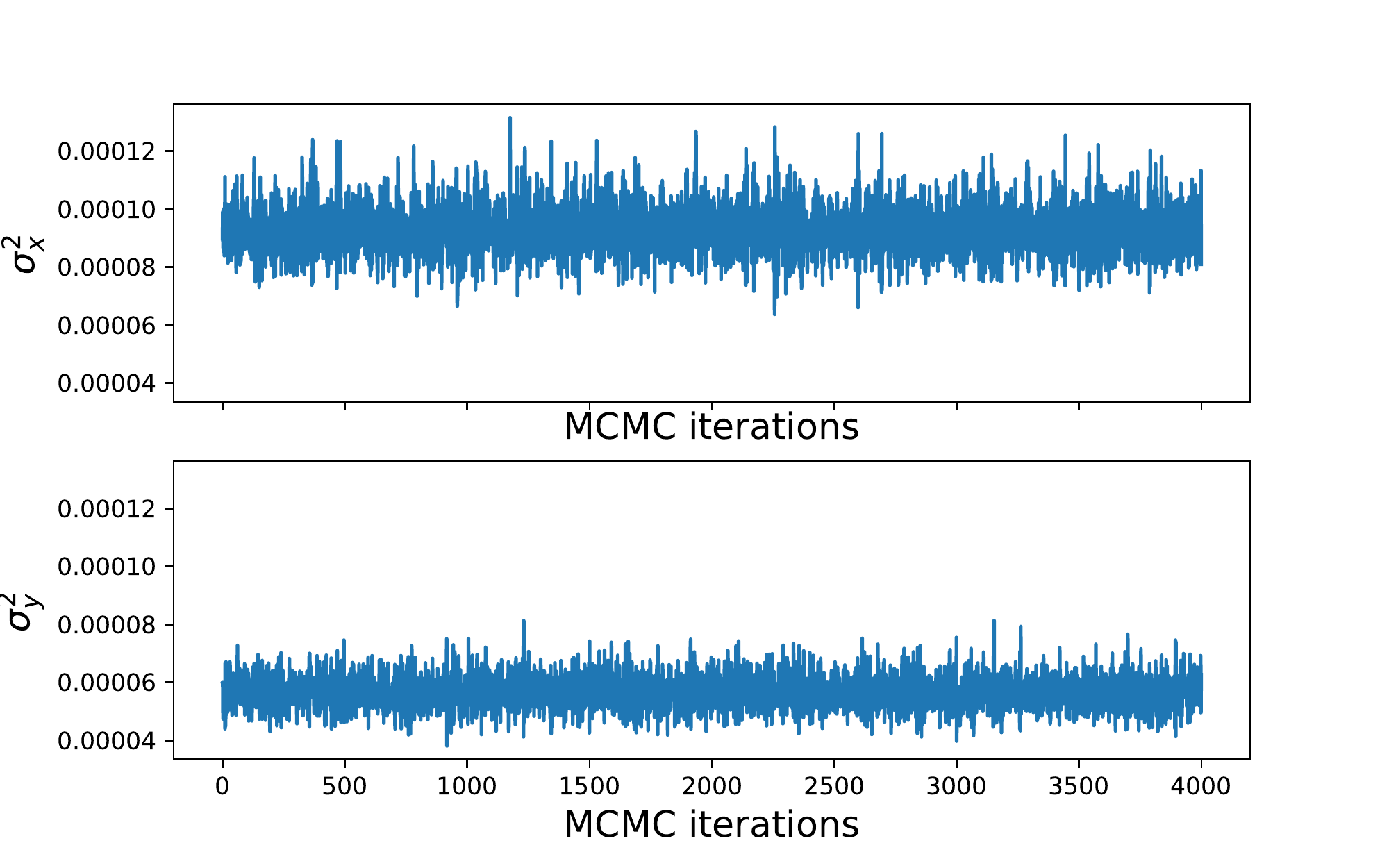}
\caption{}
\end{subfigure}
\caption{
MCMC trace for the parameters (a) $a_0,\alpha,\beta$  and (b) $\sigma_x^2,\sigma_y^2$  when collecting the MCMC samples from all chains and discarding burn-in.}%
\label{fig:trace_gunfire}
\end{figure*}

\begin{figure}[htb!]
    \centering
        \includegraphics[width=.6\linewidth, height=.3\textheight]{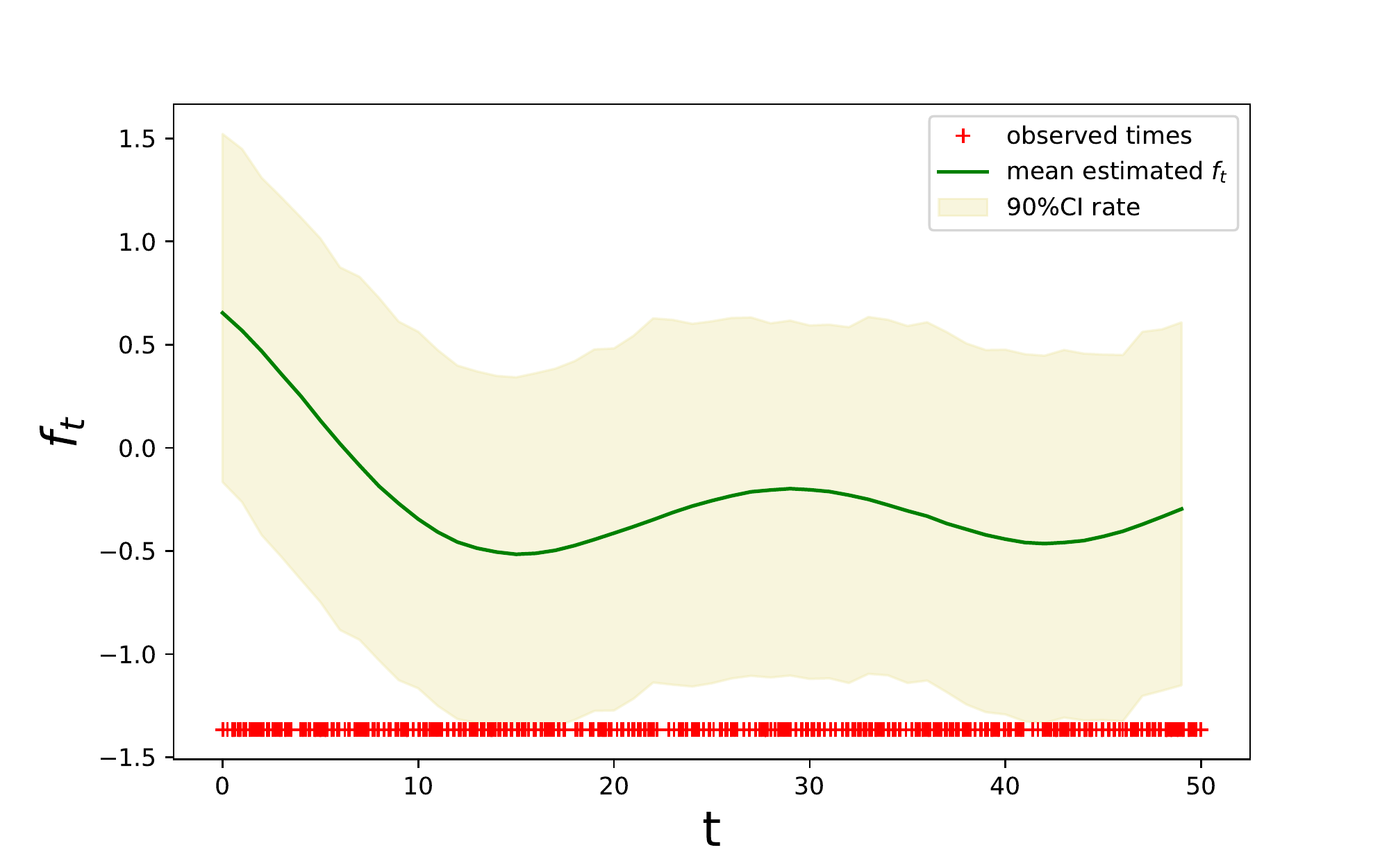}
        
        \caption{Posterior predictive for $f_t(t)$ with the posterior mean in green and the $90\%$ credible interval in the yellow shaded area, with true time stamps of the events on the x-axis.}
        
        \label{fig:post_pred_t_gunfire}
\end{figure}

\begin{figure}[htb!]
    \centering
         \includegraphics[width=.8\linewidth, height=.4\textheight]{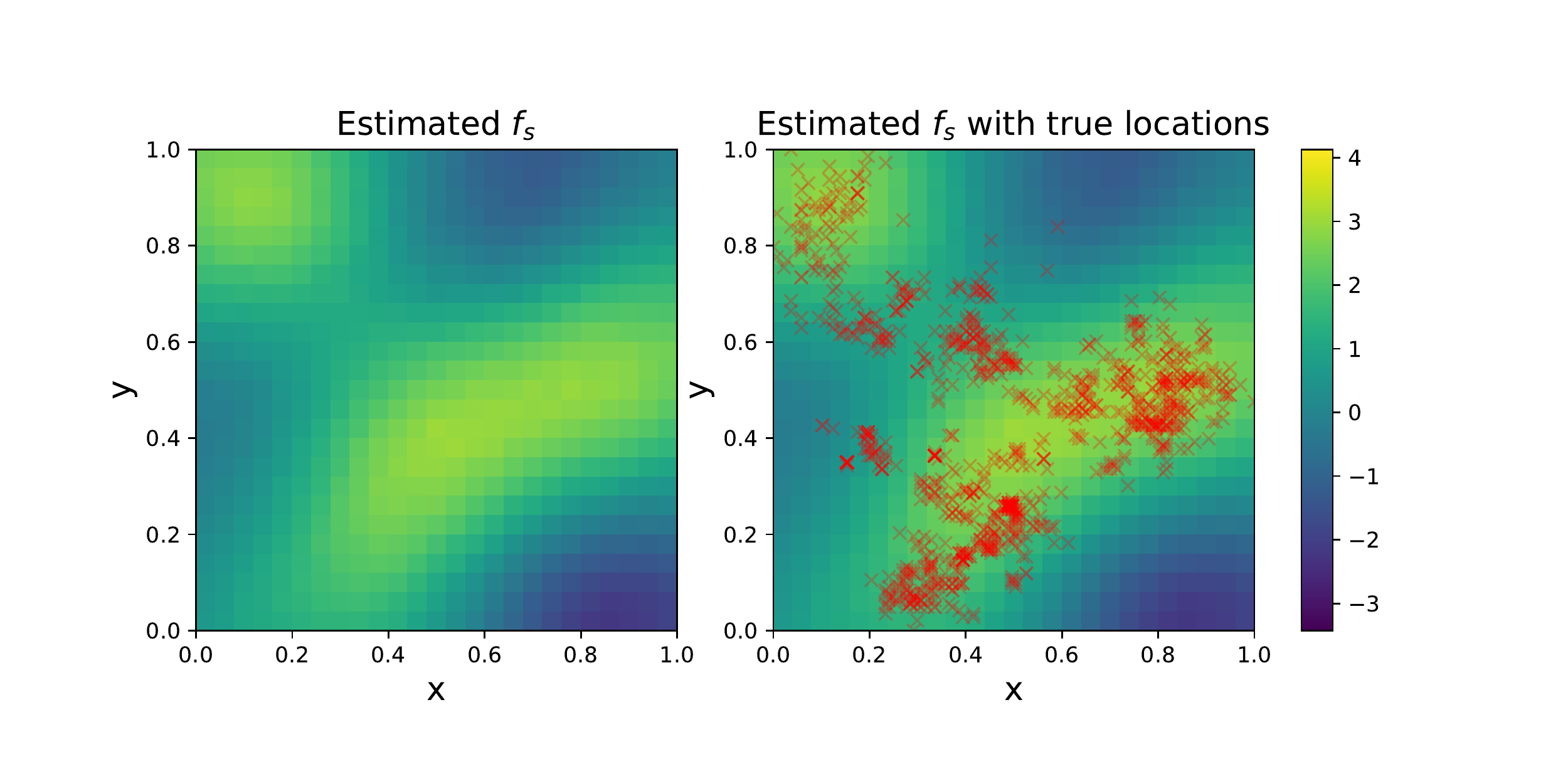}       
        \caption{Mean of the posterior predictive for $f_s(x,y)$ (left) and similarly with the true locations (right).}
        \label{fig:post_pred_s_gunfire}
\end{figure}

We perform inference with the HMC routines of numpyro, using $2$ chains each with $4,000$ samples from which $2000$ are discarded as burn-in. We join together the samples from the two chains and report the combined MCMC trace for each of the parameters. Note that we did some prior sensitivity analysis to assess the robustness of our results. We used different parameters on the priors for the parameters and we observed that the posterior distributions of the parameters were similar, giving posterior mean estimates very close to each other.
In Figure~\ref{fig:trace_gunfire}(a) we report the MCMC trace plots for the parameters $\alpha$ and $\beta$  and $a_0$ and in Figure~\ref{fig:trace_gunfire}(b) we report $\sigma^2_x, \sigma^2_y$ which define the lengthscales of the spatial Gaussian kernel (longitude and latitude) that governs excitation in space. %
Regarding the Gaussian process fitting, we show the posterior predictive plots in Figures~\ref{fig:post_pred_t_gunfire} and~\ref{fig:post_pred_s_gunfire}.
For the one-dimensional temporal Gaussian process we plot the estimated function $f_t(t)$ in Figure~\ref{fig:post_pred_t_gunfire}. The green line is the mean posterior of $f_t(t)$ and the yellow shaded area is the $90\%$ credible interval obtained from the posterior predictive distribution. The red marks indicate the observed true events. %

For the spatial Gaussian process we plot the estimated function $f_s(x,y)$ at the left plot of Figure~\ref{fig:post_pred_s_gunfire}. The mean predictive distribution with the true locations embedded on it is shown on the right.%
The plots overall show good convergence with good mixing between the chains and no multimodal behaviour. This is also quantified by the convergence diagnostics $\hat R$ which were equal to $1$ for all the parameters estimated. %

We report an estimate and $90\%$ credible intervals of
$\hat a_0= 0.53 \,(-0.46,  1.47),\, \hat \alpha =0.73 \,(0.68,0.78),\, \hat \beta= 0.18\,(0.16,0.21),\, 1/\hat \beta=5.35 (4.64,6.11),\, \sigma_x^2=9.26e^{-5}(7.90e^{-5}, 1.07e^{-4}),\,$ $\sigma_y^2=5.65e^{-5}\,(4.78e^{-5}, 6.67e^{-5})$ which can be interpreted as follows. The average number of shootings triggered by one shooting is around $0.7$. Then, rounding to the nearest minute or meter correspondingly, the temporal lengthscale for the exponential triggering kernel is estimate to be around $5$ minutes, the spatial triggering lengthscale for latitude $\sigma_x$ around $10m$ and for longitude $8m$. This means that for every 100 shootings that occur, these create at most another $73$. Using the right upper bound of the uncertainty intervals, the period in which diffusion takes place is within less than $6$ minutes and the area is within $10$ meters in $x$ distance and $8$ meters in $y$ distance. %
Regarding the background effects the posterior mean of the spatiotemporal Gaussian process is estimated to be $0.53$.
The results have some differences from the ones reported by~\cite{Flaxman2015} and~\cite{Holbrook2021} but the model assumed here has a different form and we have applied it on a different subset of the gunshot dataset.
\begin{table}[h!]
\caption{Average RMSE with its standard error in bracket computed when predicting future unseen temporal and spatial events under the four models. \label{tab:gunfire}}
\begin{center}
\begin{tabular}{rrrrrr}
            & Hawkes-LGCP &  Hawkes& LGCP & Poisson %
\\\hline
Prediction Error \vline &  7.33 (0.11) &    8.14 (0.13)  &  7.90 (0.09) &   14.2 (0.29) \\
\end{tabular}
\end{center}
\end{table}
We also compare our model to the LGCP model, Hawkes model and baseline Poisson showing how the LGCP-Hawkes gives the best predictive performance. %
We calculate the prediction error by simulating the predictions as explained before and report in Table~\ref{tab:gunfire} the mean and standard error of the RMSE in bracket. Evidently LGCP-Hawkes gives the lowest error showing how our model is generic and flexible enough to capture the underlying generating mechanism in real world scenarios.

\section{Conclusion}
\label{sec:conc}
We presented a novel model combining Hawkes processes with Gaussian processes, and used it to identify patterns in gun violence in Washington DC. Methodologically, ours is the first model of its kind to have such flexibility in capturing underlying patterns in the rate of occurrence of events, combining a powerful nonparametric statistical model with an interpretable mechanistic self-exciting point process model. This combination means that it can be used across a range of real world spatiotemporal problems in which the underlying data mechanism is unknown. Applications could include social networks, biology, economics and epidemiology. Its general and practical form make it an actionable tool for practitioners that can be used to design interventions and for policy making. 

There are many directions for future research. The model can be extended to multivariate, mutually exciting Hawkes processes, for example to model events in different regions where events in one region trigger events in other regions. This model could be useful for crime data, and also in  neuroscience, where multiple neural trains interact across different parts of the brain. Computationally, this may prove to be a difficult extension. Another interesting extension would be to consider non separable space and time. Finally, in scenarios where the background trends are potentially coming from different sources, incorporating Gaussian process mixtures would make the framework even more flexible and able to capture multimodal distributions. %

\bibliographystyle{agsm}
\bibliography{biblio}

\end{document}